\pdfoutput=1

\documentclass[11pt]{article}

\usepackage[]{emnlp2021}

\usepackage{times}
\usepackage{latexsym}
\usepackage{algorithm}
\usepackage[noend]{algpseudocode}
\usepackage{algorithmicx,algorithm}

\usepackage[T1]{fontenc}

\usepackage[utf8]{inputenc}

\usepackage{microtype}

\usepackage{graphicx}
\usepackage{booktabs,fancyvrb}
\usepackage{multirow}
\usepackage[normalem]{ulem}
\usepackage{xparse}
\usepackage{makecell}
\NewDocumentCommand\ulverb{v}{\uline{\ttfamily#1}}

\newif\ifsubmit

\ifsubmit
\newcommand{\xinyun}[1]{}
\else
\newcommand{\xinyun}[1]{\textcolor{red}{[Xinyun: #1]}}
\fi

\newcommand{\eat}[1]{}

\submitfalse

\title{Natural SQL: Making SQL Easier to Infer from Natural Language Specifications}

\author{Yujian Gan${}^{1}$ \ \ \ \ Xinyun Chen${}^{2}$ \ \ \ \ Jinxia Xie${}^{4}$ \ \ \ \  Matthew Purver${}^{1,3}$  \\ \bf  John R. Woodward${}^{1}$ \ \ \ \ John Drake${}^{5}$ \ \ \ \ Qiaofu Zhang${}^{4}$ \\

${}^{1}$Queen Mary University of London \ \ \ \  \ \ \ \ ${}^{2}$UC Berkeley \ \ \ \  \ \ \ \ 
${}^{3}$Jožef Stefan Institute  
\\
 ${}^{4}$Guangxi University of Finance and Economics \ \ \ \  \ \ \ \  ${}^{5}$University of Leicester \\
  \texttt{\{y.gan,m.purver,j.woodward\}@qmul.ac.uk}  \ \ \ \  
 \\\texttt{xinyun.chen@berkeley.edu} \ \ \ \ \texttt{john.drake@leicester.ac.uk} 
 \\ \texttt{\{jinxia\_xie,qiaofuzhang\}@hotmail.com}   

  }

\begin{document}
\maketitle
\begin{abstract}
  Addressing the mismatch between natural language descriptions and the corresponding SQL queries is a key challenge for text-to-SQL translation. 
  To bridge this gap, we propose an SQL intermediate representation (IR) called Natural SQL (NatSQL). 
  Specifically, NatSQL preserves the core functionalities of SQL, while it simplifies the queries as follows: (1) dispensing with operators and keywords such as~\textit{GROUP BY, HAVING, FROM, JOIN ON}, which are usually hard to find counterparts for in the text descriptions; (2) removing the need for nested subqueries and set operators; and (3) making schema linking easier by reducing the required number of schema items.
  On Spider, a challenging text-to-SQL benchmark that contains complex and nested SQL queries, we demonstrate that NatSQL outperforms other IRs, and significantly improves the performance of several previous SOTA models. Furthermore, for existing models that do not support executable SQL generation, NatSQL easily enables them to generate executable SQL queries, and achieves the new state-of-the-art execution accuracy \footnote{Our code and dataset are available at  \href{https://github.com/ygan/NatSQL}{https://github.com/ygan/NatSQL}.}. 
  \end{abstract}

\begin{figure*}[t]
  \includegraphics[width=0.95\textwidth]{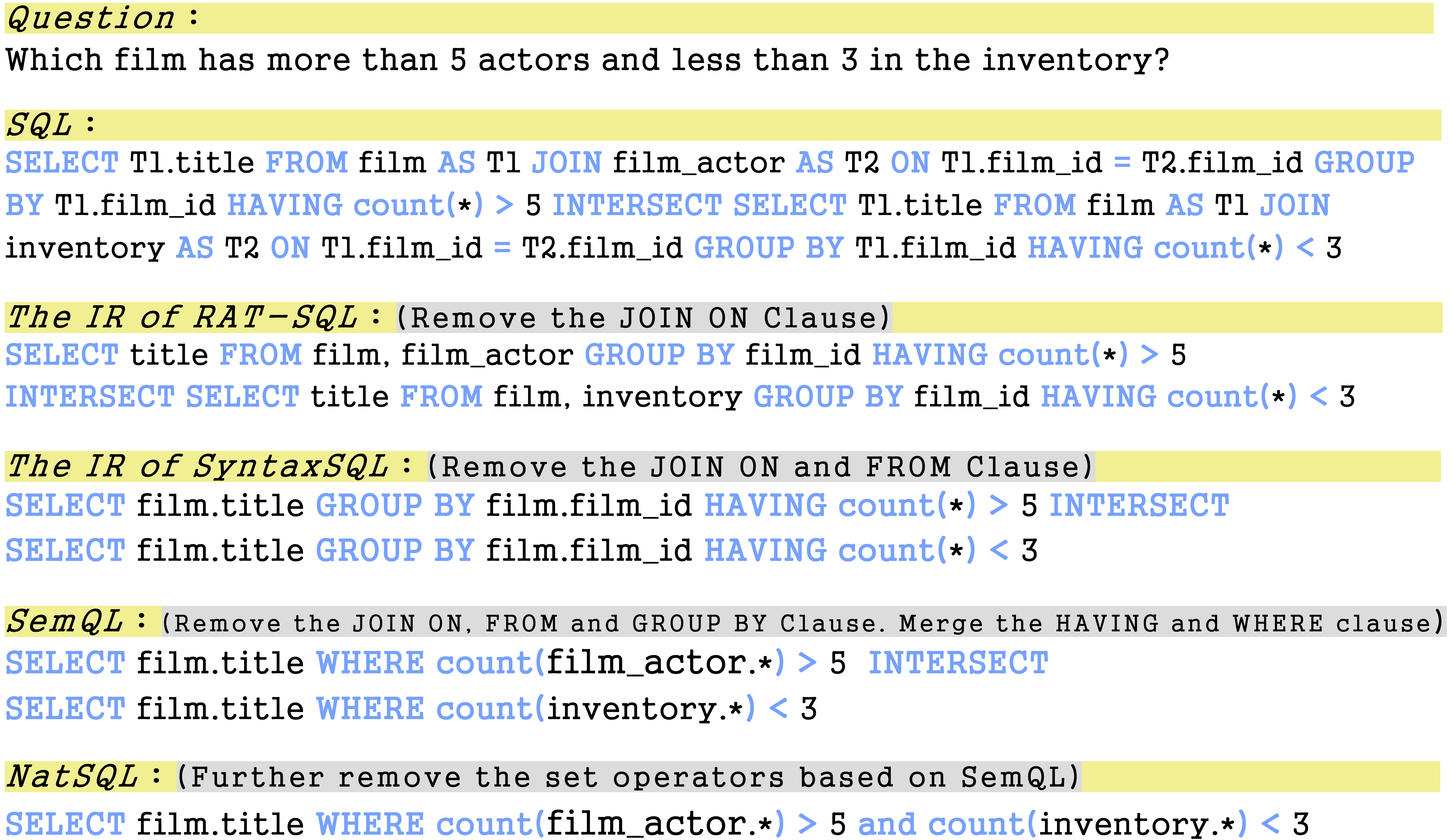}
  \centering
  \caption{A sample question in Spider dataset with corresponding SQL and IRs.}
  \label{all_ir}
\end{figure*}
\section{Introduction}
\label{section:intro}

Automatic generation of SQL queries from natural language (NL) 
has been 
studied in the literature for a number of years~\cite{Warren:1982:EEA:972942.972944,Androutsopoulos1995Natural,data-restaurants-original,Berlin2006Constructing,Dong2018,Li2014,iacob-etal-2020-neural}.
More recently, WikiSQL~\cite{zhongSeq2SQL2017}, the first large-scale cross-domain text-to-SQL dataset, has attracted much attention from the research community~\cite{Chawla2006,Wang2018rubust,x-sql}. 
Although the current state-of-the-art approach has achieved over $90\%$ execution accuracy on WikiSQL~\cite{x-sql}, since the SQL queries in this benchmark only cover a single \textit{SELECT} column and aggregation, as well as \textit{WHERE} conditions, it does not represent the true complexity of SQL generation. 
To facilitate more realistic evaluation,~\citet{Yu2018a} introduced Spider,
the first large-scale cross-domain text-to-SQL benchmark with complex and nested SQL queries, on which previous models designed for WikiSQL
suffer a significant performance drop.

To synthesize SQL queries with more complex structures, intermediate representation (IR) is widely employed by the previous SOTA models on the Spider dataset~\cite{Wang2019,Guo2019,Yu2018-SyntaxSQLNet,DBLP:journals/corr/abs-2012-10309}. 
However, previous IRs are either too complicated or have limited coverage of SQL structures. 
Besides, although the existing IRs eliminate part of the mismatch between intent expressed in NL and the implementation details in SQL, there is still some mismatch that can be further eliminated by improving the IR.

In this work, we present~\emph{Natural SQL (NatSQL)}, a new intermediate representation that offers simplified queries over other IRs, while preserving a high coverage of SQL structures. More importantly, NatSQL further eliminates the mismatch between NL and SQL, and can easily support executable SQL generation. Figure \ref{all_ir} presents a sample comparison between NatSQL and other IRs.
We observe that there is a mismatch between the NL word \verb|`and'| and the \textit{INTERSECT} SQL keyword, since in another similar question shown in Figure \ref{figure:exe-sql}, the \verb|`and'| no longer corresponds to the \textit{INTERSECT} keyword.
To translate the NL question into a corresponding query, previous IRs need the models to distinguish whether the word \verb|`and'| corresponds to \textit{INTERSECT}, this is not required for NatSQL.
Among all IRs, NatSQL provides the simplest and shortest translation, while the NatSQL structure also aligns best with the NL question.

NatSQL preserves the core functionalities of SQL, while simplifying the queries as follows: (1) dispensing with operators and keywords such as~\textit{GROUP BY, HAVING, FROM, JOIN ON}, which are usually hard to find counterparts for in the text descriptions; (2) removing the need for nested subqueries and set operators, using only one \textit{SELECT} clause in NatSQL; and (3) making schema linking easier by reducing the required number of schema items that are normally not mentioned in the NL question.
The design of NatSQL easily enables executable SQL generation, which is not naturally supported by other IRs.

We compare NatSQL with SQL and other IRs by incorporating them into existing open-source neural network models that achieve competitive performance on Spider.
Our experiments show that NatSQL boosts the performance of these existing models, and outperforms both SQL and other IRs. In particular, equipping RAT-SQL+GAP with NatSQL achieves a new state-of-the-art execution accuracy on the Spider benchmark.
These results suggest that to improve the ability of text-to-SQL models to understand and reason about the NL descriptions, designing IRs to better reveal the correspondence between natural language and query languages is a promising direction.


  \section{Review: Text-to-SQL Paradigm}
Most existing text-to-SQL models generate the SQL keywords (blue character in Figure \ref{all_ir}) and SQL schema items (black character in Figure \ref{all_ir}) separately.
Based on this paradigm, we investigate how we can design an IR to improve both SQL keyword generation and schema item generation.

\subsection{Generating SQL Keywords}
\label{sec:sql-keywords}
Neural text-to-SQL models usually generate the SQL keywords according to the similarity linking scores between the hidden state from the question and the production rule embeddings.
For example, in Figure \ref{all_ir}, we conjecture a good text-to-SQL model should be able to give a higher linking score 
between the word `\emph{less}' and the SQL `$<$' keyword.

However, SQL is designed for effectively querying relational databases, not for representing the meaning of NL questions. Hence, there inevitably exists a mismatch between intents expressed in natural language and the implementation details in SQL \cite{Guo2019}. For example, in Figure \ref{all_ir}, the \textit{GROUP BY} and \textit{JOIN ON} clauses are not mentioned in the question. 
One solution is to use an IR to remove the SQL clauses that are hard to predict.
Experiments show that the SemQL IR can improve the accuracy of previous models \cite{Guo2019}. 



\subsection{Generating Schema Items}
\label{sec:sl}


Text-to-SQL models usually generate the schema items according to the similarity linking scores between tokens in the question and database schemas. 
Intuitively, a model is supposed to predict higher scores to schema items that are mentioned in the question. 
To achieve this goal, some existing neural networks implement a schema linking mechanism, by recognizing the tables and columns mentioned in a question 
\cite{Guo2019,Bogin2019,Wang2019}.

Schema linking is essential for text-to-SQL tasks. 
As shown in the ablation study of IRNet~\cite{Guo2019} and RAT-SQL~\cite{Wang2019}, removing the schema linking results in a dramatic decrease in performance. 
%
%
The importance of schema linking raises a question about generating schema items not mentioned in the question.
Some models use graph neural networks to find these unmentioned schema items, and some models delete unmentioned schema items based on the IR; e.g., in Figure \ref{all_ir}, the IRs remove the \textit{JOIN ON} and \textit{GROUP BY} clauses with the unmentioned schema items.
  
\begin{table}[t]
  
  \centering
 
  \resizebox{.99\columnwidth}{!}{
  \smallskip\begin{tabular}{cll}
      NatSQL     & = & \textit{SELECT}  , Column ,       $\left\{ \right.$   `,' \ Column $\left. \right\}$ ,                                  \\
      &   & $\rm \left[\ \textit{WHERE}\ W\_Cond\  \right]$ , \\
                       &   & $\rm \left[\ \textit{ORDER\ BY\ } Order\_By\  \right]$ ;                                             \smallskip                  \\
      
      Column  & = & Agg\_Col\ \  $|$\ \ Table\_Col \  \ ;                                                         \smallskip              \\
      
      Agg\_Col      & = & Agg\_Fun , `('  Table\_Col , `)' ;                                  \smallskip                     \\
      
      Agg\_Fun    & = & `\textit{avg}'\ \  $|$\ \ `\textit{count}'\ \  $|$\ \ `\textit{max}'\ \  $|$\ \ `\textit{min}'\ \  $|$\ \ `\textit{sum}' ;                               \smallskip   \\
      
      Table\_Col    & = & TABLE\_NAME , `.' , COLUMN\_NAME                                                     \\
      
           &  & $|$\ \underline{TABLE\_NAME , `.' ,  $*$}  ;                                                                                  \smallskip\\
                                    
      
      W\_Cond & = &  [\underline{Conjunct}], Condition , $\left\{ \right.$   Conjunct \ Condition $\left. \right\}$ ;                                     \smallskip \\
      
      Condition        & = & Cond\_L , W\_Oper , Cond\_R ,                                                          \smallskip\\ 
                       &   & $\left[\right.$  `and' , NUMBER $\left.\right]$ ;                                                             \smallskip \\
      
      Conjunct        & = & `\textit{and}'\ \ $|$\ \ `\textit{or}'\ \ $|$\ \ `\underline{\textit{except}}'\ \ $|$\ \ `\underline{\textit{intersect}}'\ \ 
      \smallskip    \\ 
                              
                       &   & $|$\ \ \ `\underline{\textit{union}}' $|$\    `\underline{\textit{sub}}' ;                                                                                              \smallskip  \\
      
      W\_Oper  & = & `\textit{between}'\ \ $|$\ \ `='\ \ $|$\ \ `$>$'\ \  $|$\ \ `$<$'\ \ $|$\ \ `$>=$'\ \        \\
                       &   &  $|$\ \ `$<=$'              $|$\ \ `$!=$'\ \ $|$\ \ `\textit{in}'\ \ $|$\ \ `\textit{like}'\ \ $|$\ \ `\textit{is}'\ \           \\
                       &   &  $|$\ \ `\textit{exists}'\ \ $|$\ \ `\textit{not in}'  $|$\ \ `\textit{not like}'\ \
                       \\
                       &   &  
                       $|$\ \ `\textit{not between}'\ \ $|$\ \ `\textit{is not}'\ \ $|$\ \ `\underline{\textit{join}}' ;                                        \smallskip         \\
      
      Cond\_R & = & NUMBER\ \ $|$\ \ STRING\ \ $|$\ \ Column ;        \\                Cond\_L & = & Column\ \ $|$\ \ ``\underline{@}'' \ \ ;            \smallskip       \\
      
      Order\_By      & = & Column , $\left[\right.$  \textit{DESC}\ \ $|$\ \ \textit{ASC} $\left.\right]$ , \\
      & & [ \textit{LIMIT} , NUMBER ]    \\     
  \end{tabular}
  }
  \caption{The main grammar of NatSQL. Here we highlight the differences of production rules from SQL.}
  \label{table2}
  \end{table}

\section{NatSQL}

\subsection{Overview}
Table \ref{table2} presents the grammar specification of NatSQL. NatSQL only retains the \textit{SELECT}, \textit{WHERE} and \textit{ORDER BY} clauses from SQL, dispensing with other clauses such as \textit{GROUP BY, HAVING, FROM, JOIN ON}, set operators and subqueries. Tokens in capital italics are keywords of SQL and NatSQL, and other capital tokens represent special meanings, where `TABLE\_NAME' and `COLUMN\_NAME' are defined for databases, and `NUMBER' and `STRING' represent the data types.

Except for the deleted clauses, the differences between NatSQL and SQL are underlined in Table~\ref{table2}.
NatSQL implements the function of the deleted clauses by adding new keywords and allowing \emph{conjunct} to appear before the WHERE condition.
In terms of language format, NatSQL does not add new clauses, and can retain deleted clauses as needed, as in the variant NatSQL$_{\bf G}$ discussed in Section~\ref{natsql-g}. 

The main design principle of NatSQL is to simplify the structure of SQL and bring its grammar closer to natural language. 
Considering the example in Figure \ref{all_ir}, the set operator ‘\textit{INTERSECT}’, used to combine \textit{SELECT} statements, is never mentioned in the question. \textit{INTERSECT} is introduced in SQL to allow the combination of the results of multiple functions. Such implementation details, however, are rarely considered by end users and therefore rarely mentioned in questions \cite{Guo2019}. 

\subsection{Overall Comparison}

Starting from SyntaxSQLNet \cite{Yu2018-SyntaxSQLNet}, several types of IR have been developed for text-to-SQL models on the Spider dataset. The main limitation of SyntaxSQLNet is that it removes the \textit{FROM} and \textit{JOIN ON} clauses, which may result in the failure to find the correct table when converted to SQL. For example, in Figure \ref{all_ir}, SyntaxSQLNet IR misses the \emph{inventory} table, thus it cannot generate the correct \textit{JOIN ON} clause that appears in the original SQL.
The IR for RAT-SQL \cite{Wang2019} is mostly close to SQL, and it avoids missing tables since it only removes the \textit{JOIN ON} clause from SQL. 
\citet{zhong2020grounded} and \citet{lee-2019-clause} also utilize an IR that is similar to the IR in RAT-SQL and SyntaxSQLNet.

\citet{Guo2019} introduced SemQL, an intermediate language, to facilitate SQL prediction. As with NatSQL, SemQL removes the keywords \textit{FROM, JOIN ON, GROUP BY, HAVING} from SQL. Although SemQL and NatSQL remove both \textit{FROM} and \textit{JOIN ON} clauses, SemQL and NatSQL avoid missing a table by moving the table into the `*' column.
NatSQL improves on SemQL in the following ways: \\
(1) Compatible with a wider range of SQL queries than SemQL. \\
(2) Simplify the structure of queries with set operators, i.e., \textit{INTERSECT}, \textit{UNION}, and \textit{EXCEPT}, denoted as~\emph{IUE} hereafter. \\ 
(3) Eliminate nested subqueries. \\
(4) Reduce the number of schema items to predict. \\
(5) NatSQL uses the same keywords and syntax as SQL, which makes it easier to read and expand than SemQL.

There are four examples in Figure \ref{all_ir}, \ref{readability}, \ref{iue-gap} and \ref{figure0} demonstrating the differences between SQL, SemQL, and NatSQL statements representing the same natural language question.


\subsection{Scalability of NatSQL}
\label{sec:join}

\begin{figure}[t]
  \includegraphics[width=0.47\textwidth]{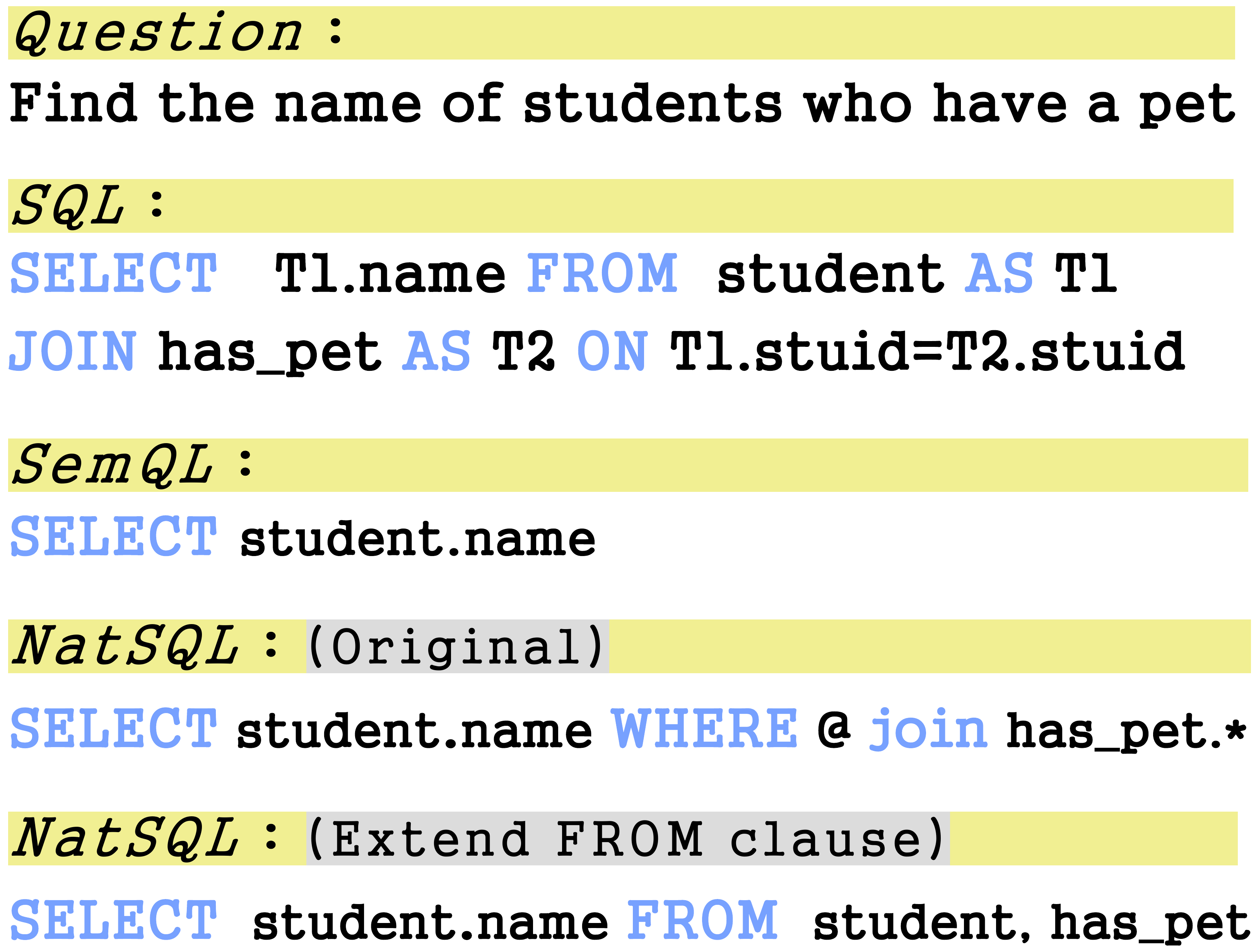}
  \centering
  \caption{An example about the scalability and readability of NatSQL.}
  \label{readability}
\end{figure}

\begin{figure}[t]
    \includegraphics[width=0.47\textwidth]{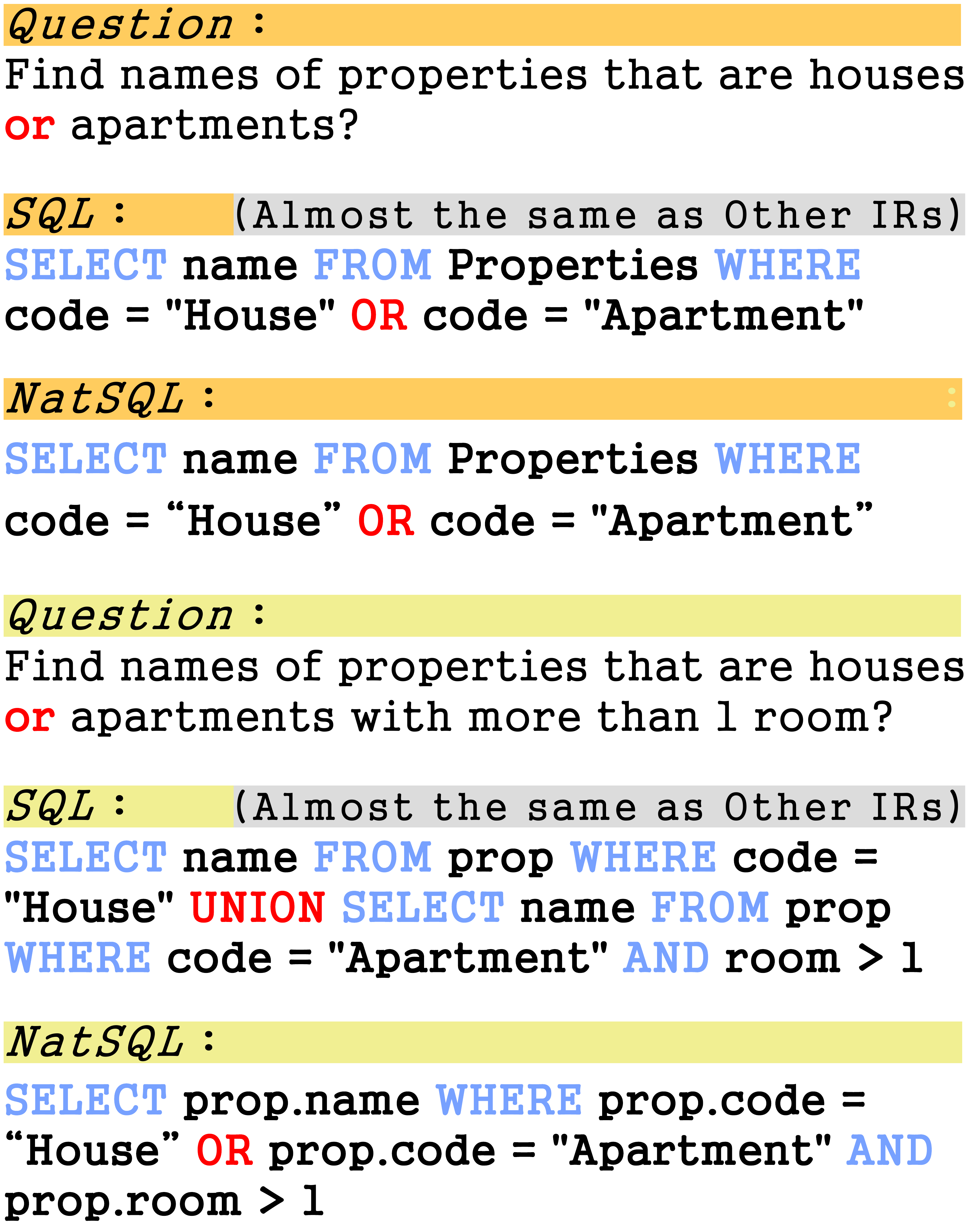}
    \centering
    \caption{An example about the mismatch between NL and IUE set operators.}
    \label{iue-gap}
  \end{figure}

We take an SQL query with multiple tables as an example. In Figure \ref{readability}, since the SemQL misses the \emph{has\_pet} table, SemQL cannot be converted to the target SQL, indicating that SemQL is not compatible with this type of SQL query. The SyntaxSQLNet IR is also not compatible, but the RAT-SQL IR can convert this query appropriately.


While both SemQL and NatSQL completely remove all \textit{FROM} and \textit{JOIN ON} clauses, NatSQL introduces a new \textit{WHERE} condition operator~\emph{join} for these unremovable \textit{JOIN ON} clauses, as shown in Figure \ref{readability}. 
With this extra \textit{WHERE} condition, NatSQL can be converted to the target SQL. 
Alternatively, you could use the NatSQL augmented with \textit{FROM} clause version.
We recommend the original version since its experimental result is better and the sub-question \verb|`who have a pet'| looks like a \textit{WHERE} condition. 
We modify this example in Table \ref{modified_ex} to illustrate why it looks like a \textit{WHERE} condition.
Usually, NatSQL does not need the~\emph{join} operator for generating \textit{JOIN ON} clause, such as the `\emph{Ques 2}' in Table \ref{modified_ex}, except in cases when it cannot infer the correct \textit{JOIN ON} clause from other clauses.

\paragraph{NatSQL$_{\bf G}$.}  
Since each database has different compatibility with SQL, we allow NatSQL to retain the deleted clauses as needed.
NatSQL$_{\bf G}$ is NatSQL augmented with \textit{GROUP BY}, which improves the compatibility in the SQLite database where the Spider benchmark is built on.
\label{natsql-g}

\begin{table}[t]
  \centering
 
  \resizebox{.99\columnwidth}{!}{
  \smallskip\begin{tabular}{cl}
    Ques 1:& Find ... who have a pet. \\\hline

	NatSQL:    &  ... {\bf WHERE} @ \textit{join}  has\_pet.* \\\hline
    
  Ques 2:& Find ... who have two pet. \\\hline

  NatSQL:    &   ... {\bf WHERE} \textit{count}(has\_pet.*) = 2 \\\hline
  
  \end{tabular}
  }
  \caption{A modified example based on Figure \ref{readability}}
  \label{modified_ex}
  \end{table}

\subsection{NatSQL for SQL Keyword Generation}

By simplifying the set operators and nested subqueries, NatSQL improves text-to-SQL models.

\subsubsection{Simplifying Queries with Set Operators}

It is typically hard to generate queries with IUE (\textit{INTERSECT, UNION}, and \textit{EXCEPT}) set operators for text-to-SQL models, where the corresponding F1 score is usually the lowest among all breakdown metrics on the Spider benchmark~\cite{Guo2019,Bogin2019,Wang2019}. The main reason is that the related questions are generally longer and more complicated, while the mismatch between NL and SQL queries further increases the prediction difficulty, as discussed in Section \ref{sec:sql-keywords}.

Figure \ref{iue-gap} compares the SQL queries corresponding to two similar problems. 
The second question in Figure \ref{iue-gap} contains an extra condition: \verb|`more than 1 room'|.
This extra condition changes the structure of the entire SQL query. 
Although IRs have been widely used for complex SQL, enthusiasts of end-to-end models expect the text-to-SQL model to automatically distinguish whether the word token \verb|`or'| in Figure \ref{iue-gap} corresponds to \textit{UNION} or \textit{OR} keyword.
However, most models cannot do that and would generate a \textit{OR} clause for both questions.
This example is similar to the comparison between Figure \ref{all_ir} and Figure \ref{figure:exe-sql} discussed on Section \ref{section:intro}.

NatSQL bridges this gap by unifying them into a simple \textit{OR} operator that will be converted to a \textit{UNION} clause when it cannot concatenate its following conditions.
The reasons for the failure to concatenate conditions include: (1) the precedence of the following conditions is higher (e.g., the precedence of \textit{AND} is higher than \textit{OR}); (2) the two conditions cannot be connected, or they are disjoint such as the example in Figure \ref{all_ir}. 
The `\emph{count(film\_actor.*)>5}' condition cannot be connected with the `\emph{count(inventory.*)<3}' condition because they belong to different tables. 
Based on the same rules, NatSQL simplifies the SQL with other set operators, the details of which can be found in Appendix A.

\subsubsection{Eliminating Nested Subqueries}

\begin{figure}[t]
  \includegraphics[width=0.47\textwidth]{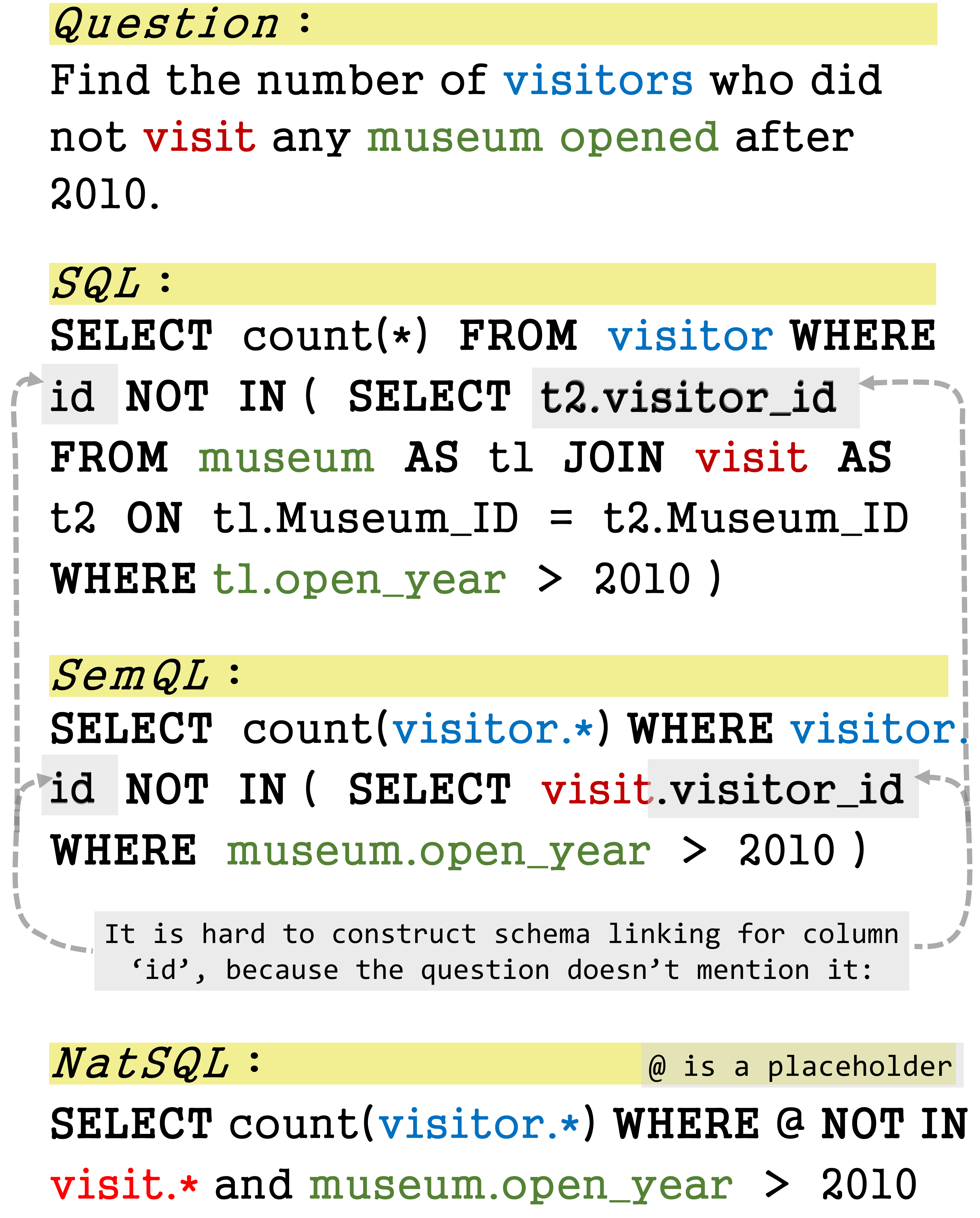}
  \centering
  \caption{A sample question in Spider dataset with corresponding SQL, SemQL and NatSQL queries.}
  \label{figure0}
\end{figure}

Since the subqueries in both NatSQL and SemQL only appear in \textit{WHERE} conditions, only one column in the \textit{SELECT} clause of a subquery is required. 
NatSQL keeps this \textit{SELECT} column in `\emph{Cond\_R}' (right column of \textit{WHERE} conditions) instead of a whole \textit{SELECT} clause. Since this meets the \textit{WHERE} condition format,  NatSQL can remove the brackets and subqueries from SQL, as shown in Figure \ref{figure0}.

\subsection{How NatSQL Help Schema Item Generation}

\begin{algorithm*}[t]
  \caption{Infer columns to replace the @ and table.* in NatSQL}\label{alg:infer}
  \hspace*{0.02in} {\bf Input:} 
  $t\_list$ \Comment{All tables before @, which include the table `visitor' in Figure \ref{figure0}}\\
  \hspace*{0.52in}$table\_r$ \Comment{The table next to the @, which is the table `visit' in Figure \ref{figure0} }
  \\
  \hspace*{0.02in} {\bf Output:} 
  Two columns to replace the @ and table.*
  \begin{algorithmic}[1]
  \For {Every $table$ in $t\_list$}
    \If {There is foreign key relationship between $table$ and $table\_r$ }
    \State \textbf{return} These two foreign key columns
  \EndIf
  \EndFor
  
  \For {Every $table$ in $t\_list$}
  \If {There are columns with the same name in both $table$ and $table\_r$ }
  \State \textbf{return} The same name columns
  \EndIf
  \EndFor
  \State \textbf{return} Their primary keys
  \end{algorithmic}
  \end{algorithm*}

NatSQL helps schema item generation by reducing the number of schema items that need to be predicted.
For example, in Figure \ref{figure0}, without an in-depth analysis of the database schema, by looking at the natural language description itself, it is difficult to infer the grey shaded columns in SQL and SemQL (in this example, they are column `\emph{id}' in table `\emph{visitor}' and column `\emph{visitor\_id}' in table `\emph{visit}'). 
We cannot build a schema linking for these columns, even though the schema linking is important to boost performance as discussed in Section~\ref{sec:sl}.

NatSQL solves this problem by replacing some of the columns with a table only or @, where @ is a place holder of NatSQL.
We can find that all columns of NatSQL in Figure \ref{figure0} are mentioned in the question. Specifically, NatSQL uses @ to replace the `\emph{visitor.id}' and uses `\emph{visit.*}' to replace `\emph{visit.visitor\_id}'.


@ is a placeholder in NatSQL that only appears in `\emph{Cond\_L}', which denotes that we need to infer a column to replace it.
The `*' keyword does not appear in the \textit{WHERE} condition without an aggregation function, so NatSQL uses it to represent a table. With this table, we can infer the correct column in the target SQL to replace the @ and `\emph{table.*}' according to Algorithm~\ref{alg:infer}.





\begin{figure*}[t]
  \includegraphics[width=0.95\textwidth]{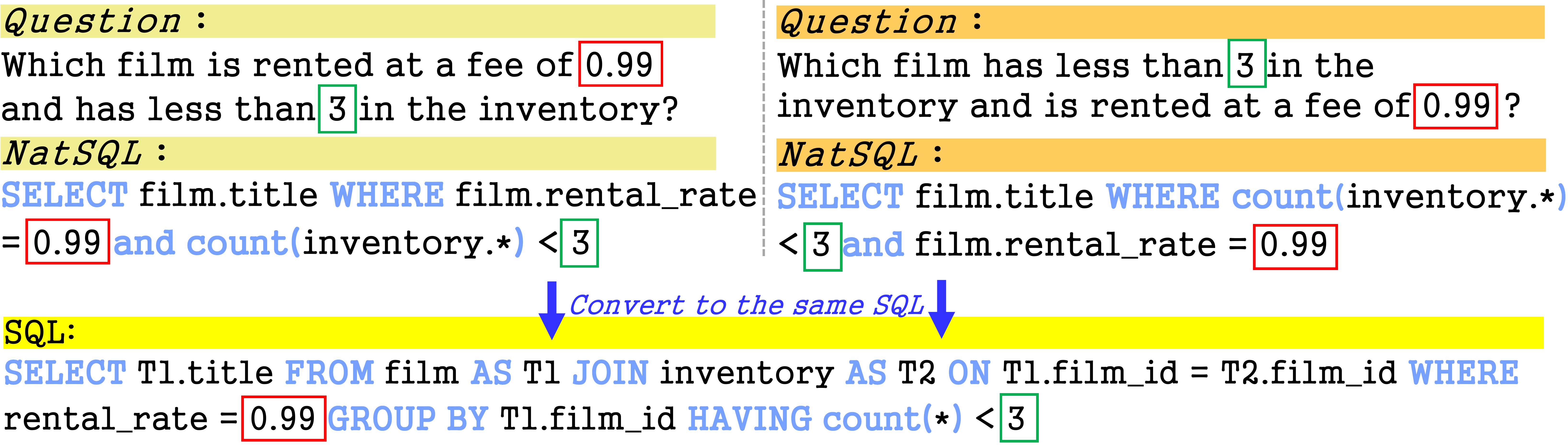}
  \centering
  \caption{Fill the values in order of appearance (see more discussion in Appendix B).}
  \label{figure:exe-sql}
\end{figure*}

\subsection{Executable SQL Generation}
\label{sec:exec-gen}
Many previous text-to-SQL models \cite{Guo2019,Wang2019,Bogin2019} only focus on the Spider exact match accuracy, i.e., they only generate the SQL queries without condition values.
These queries are not executable until filling in the condition values.
However, it is not easy to fill in the values correctly.
On the one hand, there are too many possible condition value slots that need to be searched. The slots can appear in:  \textit{WHERE} clause,  \textit{WHERE} clause in a subquery,  \textit{WHERE} clause after set operators, \textit{HAVING} clause, 
etc.
On the other hand, when there are multiple value slots, it is easier to confuse where to fill.
For example, in Figure \ref{figure:exe-sql}, the two different questions correspond to the same SQL query, making it hard to copy the right values from the question to SQL.

Because the condition value slots of NatSQL only appear in the \textit{WHERE} clause, generating condition values becomes much easier, as shown in Figure~\ref{figure:exe-sql}.
Unlike the models \cite{lin-etal-2020-bridging,DBLP:journals/corr/abs-2010-12412} trained to copy the values from questions to SQL queries, NatSQL simply copies the possible values (numbers or database cell values) from questions to SQL in the order of appearance without training.
This feature enables the models designed only for the Spider exact match metrics to generate executable SQL.
  \section{Experiments}
\label{sec:eval}

\subsection{Experimental Setup}
We evaluate NatSQL on the Spider benchmark \cite{Yu2018a}. There are 7000, 1034 and 2147 samples for training, development and testing respectively, where 206 databases are split into 146 for training, 20 for development and 40 for testing.

We first evaluate the gold NatSQL and other IRs using the exact match and execution match metrics in~\cite{Yu2018a}. 
Exact match measures whether the predicted query without condition values as a whole is equivalent to the gold query.
Execution match measures whether the execution result of the predicted query from the database is the same as the gold query.
We then evaluate NatSQL and other IRs using existing open-source models that provide competitive performance on Spider: (1) GNN \cite{Bogin2019}; (2) IRNet \cite{Guo2019}; (3) RAT-SQL \cite{Wang2019}; (4) RAT-SQL+GAP \cite{DBLP:journals/corr/abs-2012-10309}. 
Although some of these models are not designed for the generation of executable SQL queries, with the approach discussed in Section~\ref{sec:exec-gen}, we utilize NatSQL to generate executable SQL and evaluate the execution match performance.

\subsection{Comparison Between IRs}
\subsubsection{Gold IRs}
\label{Gold IRs}
In Table~\ref{table3}, we present the exact match and execution match accuracies of the gold IRs on the Spider development set, where the metrics are defined by \citet{Yu2018a} for the Spider benchmark.
\begin{table}[h]
    \centering
    \resizebox{.99\columnwidth}{!}{
    \smallskip\begin{tabular}{ccc}
    \hline
      Language & \ \ Exact Match \ \ & Execution Match
      \\ 
      \hline
      \hline
      SQL & 100\% & 100\% \\
      \hline
      SemQL & 86.2\% & Unsupported \\
      IR(RAT-SQL) & {97.7\%} &  97.1\%
      \\ 
      NatSQL & {93.3\%} & {95.3\%}
      \\
      NatSQL$_{\bf G}$ & {96.2\%} & {96.5\%}
      \\ \hline
    \end{tabular}
    }
    \caption{The comparison between gold IRs on Spider development set.}
\label{table3}
\end{table}

\begin{table}[t]
  \centering
 
  \resizebox{.99\columnwidth}{!}{
  \smallskip\begin{tabular}{cl}
  Ques:& Find students whose age is 10 or 16. \\\hline \hline
    SQL 1:& ... {\bf WHERE} age = 10 {\bf or} age = 16  \\\hline
	NatSQL 1:    & ... {\bf WHERE} age = 10 {\bf or} age = 16 \\\hline \hline
    
  SQL 2:& ... {\bf WHERE} age = 10  {\bf UNION}
  \\ &  ... {\bf WHERE} age = 16  \\
  \hline
	NatSQL 2:    & ... {\bf WHERE} age = 10 {\bf union} age = 16 \\\hline
  
  \end{tabular}
  }
  \caption{Equivalent SQL queries with its NatSQL}
  \label{Equivalent SQL queries}
  \end{table}
  
We observe that NatSQL can be converted to more gold SQL than SemQL, because NatSQL can handle the unremovable \textit{JOIN ON} clauses, as discussed in Section~\ref{sec:join}. 
Such SQL queries comprise around 5\% of the entire Spider dataset. 
Other performance improvement comes from the fact that NatSQL is more compatible with subqueries and that its capability to generate SQL is better. 
More importantly, SemQL is designed only for the exact match metrics of Spider, and cannot directly be used to generate executable SQL.

The IR of RAT-SQL is the most similar to SQL and thus has the highest coverage among all IRs. However, NatSQL$_{\bf G}$ further simplifies the queries with only 0.6\% execution accuracy degradation, whilst enabling better model prediction performance.
NatSQL$_{\bf G}$ outperforms NatSQL when comparing the gold queries, but the gap is small when they are utilized by models. We defer more breakdown analysis to Appendix~C.

The result in the training set is close to that in the development set. 
It should be noted that the exact match accuracy will slightly vary in different NatSQL versions. 
The accuracy depends on the attitude towards equivalent SQL queries. 
Table~\ref{Equivalent SQL queries} presents two equivalent SQL queries with their corresponding NatSQL queries. 
Considering that \textit{UNION} is not mentioned in the question, we prefer to sacrifice the exact match accuracy for a more succinct NatSQL representation, i.e., we will use the first NatSQL query in Table~\ref{Equivalent SQL queries} to represent the second SQL, even though it can not be converted into the second SQL query.
Although our preference slightly affects the exact match accuracy in the Spider benchmark, it brings greater potential and convenience when outside Spider.

\begin{table}[t]
    \centering
    \resizebox{.99\columnwidth}{!}{
    \smallskip\begin{tabular}{lcc}
        \hline
        Approach & Exact & Execution\\
        \hline \hline
        GNN + SQL\ \  \ \ \ \ \ \ \ \ \ \ \   \ \ \ \ \ \ \ \ \ \ \ \ \  & 47.5\%   \\ 
        GNN + SemQL\ \  \ \ \ \ \ \ \ \ \ \ \   \ \ \ \ \ \ \ \ \ \ \ \ \  & 51.6\%   \\ 
        GNN + NatSQL & \bf 53.8\% & \bf 58.0\% \\ 
        \hline
        IRNet + SemQL & 51.8\%  \\ 
        IRNet + NatSQL &  \bf 52.9\% & \bf 52.6\% \\ 
        \hline
        RAT-SQL + IR(RAT-SQL) &  62.7\% \\
        RAT-SQL + SemQL & 58.4\%  \\ 
        RAT-SQL + NatSQL &   64.4\% & 66.7\%\\ 
        RAT-SQL + NatSQL$_{\bf G}$ &  \bf 65.2\% & \bf 67.3\%\\ 
        \hline
        extend BERT: \\
        RAT-SQL + IR(RAT-SQL) &  69.5\% \\
        RAT-SQL + NatSQL  &   71.7\% &  72.8\%\\ 
        RAT-SQL + NatSQL$_{\bf G}$  &  \bf 72.1\% & \bf 73.0\%\\ 
        \hline 
        extend GAP: \\
        RAT-SQL + IR(RAT-SQL)  &  71.8\% \\
        RAT-SQL + NatSQL  &\bf 73.7\% &  74.6\% \\ 
        RAT-SQL + NatSQL$_{\bf G}$  &\bf73.7\% & \bf 75.0\% \\ 
        \hline 
    \end{tabular}
    }
    \caption{Exact and execution match accuracy on Spider development set.}
    \label{table7}

\end{table}

\subsubsection{IRs for Prediction}
Table \ref{table7} presents the exact match accuracy of four models with SemQL, its default IR (or SQL), and NatSQL separately. 
We observe that NatSQL consistently outperforms SemQL with all of these model architectures, including IRNet. 
Note that the original Spider dataset additionally includes 1,659 training samples from 6 earlier text-to-SQL benchmarks (Academic, GeoQuery, IMDB, Restaurants, Scholar and Yelp), which were used to train models with SemQL in the IRNet. 
To provide a fair comparison with other baselines, we didn’t include these additional samples for all models in our evaluation, thus our presented result for IRNet+SemQL (51.8\%) is lower than the number reported in the IRNet paper (53.2\%). 

Note that SemQL causes performance decline for RAT-SQL. We hypothesize that this is because the exact match accuracy of the gold SemQL is only 86.2\%. 
With the improvement of model architectures, such a gap will affect the prediction accuracy more negatively. 
Although the accuracy of gold RAT-SQL IR is higher than that of NatSQL, NatSQL still outperforms the original RAT-SQL model, and NatSQL$_{\bf G}$ slightly improves the performance over NatSQL.


Meanwhile, NatSQL helps these models generate executable SQL queries. Execution match accuracy improves with the improvement of the exact match, and most execution match accuracy is better than that of exact match.
The execution match accuracy of IRNet is slightly lower than the exact match, because the IRNet does not predict the \textit{DISTINCT} keyword while the exact match metric does not check this aspect.

\paragraph{Breakdown results.}
Based on the complexity of the SQL, the examples in Spider are classified into four types: \verb|easy, medium, hard,| and \verb|extra hard|. 
We provide a breakdown comparison on the Spider development set, as shown in Table \ref{table:breakdown}.
The improvement brought by NatSQL mainly comes from the \verb|extra hard| SQL, which demonstrate an average 4.74\% absolute improvement across these models.
This improvement is in line with the design of NatSQL, i.e., most \verb|extra hard| SQL queries contain set operators or subqueries, while NatSQL has simplified these components.
Since \verb|easy| and \verb|medium| SQL queries categorized in the Spider dataset are more similar to NatSQL queries, it is expected that the improvement on simple SQL is less significant. 
However, we still observe that NatSQL consistently increases the accuracy on most samples of different difficulty levels.

\begin{table}[t]
    \centering
    \resizebox{.99\columnwidth}{!}{
    \smallskip\begin{tabular}{lcccc}
        \hline
        Approach & Easy & Medium & Hard & Extra\\
        \hline \hline
        GNN + SemQL\ \  \ \ \ \ \ \ \ \ \ \ \   \ \ \ \ \ \ \ \ \ \ \ \ \  & 68.5\%  & \bf 58.9\% & 36.8\% & 24.1\% \\ 
        GNN + NatSQL & \bf 72.0\% & 58.0\% & \bf 42.0\% & \bf 28.2\% \\ 
        \hline
        IRNet + SemQL & 69.8\% & 53.0\% & \bf 46.0\% & 30.1\% \\ 
        IRNet + NatSQL &  \bf 70.6\% & \bf 54.1\% & \bf 46.0\% & \bf 32.5\% \\
        \hline
        RAT-SQL + IR(RAT-SQL) &  80.4\% & 63.9\% & 55.7\% & 40.6\%\\
        RAT-SQL + NatSQL$_{\bf G}$ &  \bf 82.4\% & \bf 65.0\% & \bf 59.2\% & \bf 46.5\%\\ 
        \hline
        extend BERT: \\
        RAT-SQL + IR(RAT-SQL) &  86.4\%  & 73.6\% & 62.1\% & 42.9\% \\
        RAT-SQL + NatSQL$_{\bf G}$  &\bf 88.4\% & \bf 76.6\% &\bf 62.6\% & \bf 46.4\% \\
        \hline 
        extend GAP: \\
        RAT-SQL + IR(RAT-SQL)  &  88.3\% & 74.0\% & 64.4\%  & 44.0\% \\
        RAT-SQL + NatSQL$_{\bf G}$  &\bf91.6\% & \bf 75.2\% &\bf 65.5\% & \bf 51.8\% \\ 
        \hline 
    \end{tabular}
    }
    \caption{Exact match accuracy by difficulty on Spider development set.}
    \label{table:breakdown}

\end{table}




\subsection{Overall Performance Analysis}
First, we present the exact and execution match accuracy of our approach applied to RAT-SQL augmented with GAP in Table \ref{table-exp-result}, where we compare with various baselines at the top of the Spider leaderboard.
By incorporating NatSQL into the RAT-SQL model with GAP, we demonstrate that our approach achieves a new state-of-the-art on Spider execution benchmark, surpassing its best counterparts by 2.2\% absolute improvement.

Considering that the gap between dev and test in exact match is larger than that in execution match, we speculate that there are two reasons why our exact match accuracy has dropped by 1\% compared to RAT-SQL+GAP. 
From the complexity breakdown accuracy between dev and test, we observe that the main performance degradation comes from the \verb|extra hard| SQL queries. 
Since there are many subqueries in \verb|extra hard| SQL queries, some limitations of the Spider exact match evaluation process (discussed in Appendix~C) may have a negative effect on our prediction results.
On the other hand, some degradation may come from equivalent SQL queries. As we discuss in Section~\ref{Gold IRs} and Table~\ref{Equivalent SQL queries}, it is not mandatory to keep the NatSQL queries consistent with the original SQL queries.
As a result, the model trained by NatSQL may output equivalent SQL queries that do not match exactly but that get the same query result. 
Therefore, our evaluation shows that NatSQL is more suitable for generating executable SQL queries.

\begin{table}[t]
    \centering
    \resizebox{.99\columnwidth}{!}{
    \smallskip\begin{tabular}{llc}
        \hline
       \bf Approach & \bf Exact & \bf Execution \\
        \hline \hline
        IRNet + BERT  \cite{Guo2019} & 54.7\% & -- \\ 
        RATSQL + BERT \cite{Wang2019} &65.6\% &  -- \\ 
        BRIDGE v2 + BERT(ensemble) \cite{lin-etal-2020-bridging} & 67.5\% & 68.3\% \\ 
        COMBINE (Anonymous)   & 67.7\% & 68.2\% \\ 
        SmBoP + GraPPa \cite{DBLP:journals/corr/abs-2010-12412} & 69.5\% &  71.1\% \\ 
        RATSQL + GAP \cite{DBLP:journals/corr/abs-2012-10309} & 69.7\% & -- \\ 
        DT-Fixup SQL-SP + RoBERTa (Anonymous)& \bf 70.9\% & -- \\ 

        \hline 
        {\bf RAT-SQL + GAP + NatSQL$_{\bf G}$} (Ours)\ \  &  68.7\% & \bf 73.3\% \\ 
        \hline 
    \end{tabular}
    }
    \caption{Results on Spider test set, compared to other models at the top of the leaderboard.}\smallskip
    \label{table-exp-result}
     
\end{table}

  \section{Related Work}

\paragraph{Natural Language Interface to Database} The study of Natural Language Interface to Database (NLIDB) has a long history that can be traced back to the 1970s \cite{Warren:1982:EEA:972942.972944,Androutsopoulos1995Natural,Popescu2004,Berlin2006Constructing,iacob-etal-2020-neural}. 
Most of the early work focuses on single-domain datasets, including ATIS, GeoQuery \cite{data-atis-geography-scholar}, Restaurants \cite{data-restaurants-original,data-restaurants-logic,data-restaurants}, Scholar \cite{data-atis-geography-scholar}, Academic \cite{Li2014}, Yelp and IMDB \cite{data-sql-imdb-yelp} and so on. \citet{Finegan-Dollak2018} shows some models dealing with specific databases that only learn to match semantic parsing results. It is a challenge to generate SQL queries in a cross-domain setting, such as the case of the WikiSQL \cite{zhongSeq2SQL2017} and Spider \cite{Yu2018a} benchmarks. SyntaxSQLNet \cite{Yu2018-SyntaxSQLNet} was the first study to use the Spider benchmark. Following this work, 
many models are presented to address this problem \cite{Bogin2019,Guo2019,Zhang2019,Bogin2019a,Wang2019,DBLP:journals/corr/abs-2010-12412,lin-etal-2020-bridging}.

\paragraph{Intermediate Representations for NLIDB} 
Early work on IR of SQL tried to use an IR to translate a natural language question and then convert it to SQL queries \cite{Woods1978,Li2014}. 
\citet{LiPan2014} proposed an IR for SQL called Schema-free SQL, for users who do not need to know all of the schema information. 
The IR in SyntaxSQLNet \cite{Yu2018-SyntaxSQLNet} represents an SQL statement without \textit{FROM} and \textit{JOIN ON} clauses. 
SemQL \cite{Guo2019} removes the \textit{FROM, JOIN ON} and \textit{GROUP BY} clauses, and combines the \textit{WHERE} and \textit{HAVING} conditions. 
The IR in EditSQL \cite{Zhang2019} also combines the \textit{WHERE} and \textit{HAVING} conditions but keeps the \textit{GROUP BY} clause. 
IR is also used to improve compositional generalization in semantic parsing \cite{DBLP:journals/corr/abs-2104-07478}.
Compared to existing IRs for SQL, our NatSQL further simplifies the SQL language, moving closer towards bridging the gap between natural language descriptions and SQL statements.
  \section{Conclusion}
In this paper, we propose NatSQL, a new SQL intermediate representation that reduces the difficulty of schema linking and simplifies the SQL structure.
By incorporating NatSQL into existing neural models for text-to-SQL generation, we show that NatSQL is easier to infer from natural language specification than the full-fledged SQL and other intermediate representation languages. 
Furthermore, NatSQL enables existing models to easily generate executable SQL queries without modifying their architecture.
Experimental results on the challenging Spider benchmark demonstrate that NatSQL consistently improves the prediction performance of several neural network architectures and achieves the state-of-the-art, showing the effectiveness of our approach.


\section*{Acknowledgements}
We thank Tao Yu, Yusen Zhang and Bo Pang for  their careful assistance with the evaluation. We also thank the anonymous reviewers for their helpful comments.
Matthew Purver is partially supported by the EPSRC under grant EP/S033564/1, and by the European Union's Horizon 2020 programme under grant agreement 
825153 (EMBEDDIA, Cross-Lingual Embeddings for Less-Represented Languages in European News Media). Xinyun Chen is supported by the Facebook Fellowship. The results of this publication reflect only the authors' views and the Commission is not responsible for any use that may be made of the information it contains.

\bibliographystyle{acl_natbib}
\bibliography{emnlp2021}

\appendix

\section{Further Discussion on Set Operators}

Based on the rules discussed on Section 3.4.1, NatSQL can simplify the SQL with \textit{INTERSECT} (example is shown in Figure 1) and \textit{EXCEPT}. 
As to the case that the set operator itself represents part of a condition, NatSQL allows them to follow the \textit{WHERE} keyword. 
As illustrated in Table \ref{where-except}, this type of SQL is mainly related to the \textit{EXCEPT} operator.

The NatSQL prediction work in Table \ref{where-except} is easier than others.
NatSQL here only needs to predict the `\emph{cartoon}' table, instead of predicting the `\emph{cartoon.channel}' column. 
Predicting a table is easier than predicting a column because the premise of finding the correct column is to find the correct table.
Besides, many models incorrectly output `\emph{cartoon.id}' instead of `\emph{cartoon.channel}' because the annotation of `\emph{cartoon.id}' is the same as`\emph{tv\_channel.id}' column.

\begin{table}[h]
  \centering
 
  \resizebox{.99\columnwidth}{!}{
  \smallskip\begin{tabular}{cl}

    Ques & Find the id of tv channels that do not play any cartoon\\
    \hline
    SQL & {\bf SELECT} id {\bf FROM} tv\_channel {\bf EXCEPT} \\ &{\bf SELECT} channel {\bf FROM} cartoon
    \\ 
    \hline
    SemQL & {\bf SELECT} tv\_channel.id  {\bf EXCEPT} \\ & {\bf SELECT} cartoon.channel \\
    \hline
    NatSQL & {\bf SELECT} tv\_channel.id  {\bf WHERE} {\bf except} cartoon.* \\
    \hline

  \end{tabular}
  }
  \caption{An example of none \textit{WHERE} conditions before the IUE.}
  \label{where-except}
  \end{table}


In addition to the conditions mentioned in Section 3.4.1 that cannot be concatenated, Table \ref{figure2union} present one more example. 
These two conditions can not concatenate because one \textit{WHERE} condition can not concatenate a \textit{HAVING} condition by a \textit{OR} operator.

\begin{table}[h]
  \centering
 
  \resizebox{.99\columnwidth}{!}{
  \smallskip\begin{tabular}{cl}

    \\
    Ques & Which film is rented at a fee of 0.99 {\bf {\color{red}or}}  has less  \\ &   than 3 in the  inventory?\\
    \hline
    SemQL & {\bf SELECT} film.title {\bf WHERE} film.rental\_rate = 0.99  \\ &  {\bf {\color{red}UNION} }    \\ & 
    {\bf SELECT} film.title {\bf WHERE} count(inventory.*)$<$ 3
    \\ \hline
    NatSQL & {\bf SELECT} film.title {\bf WHERE} film.rental\_rate = 0.99 \\ &
    {\bf {\color{red}OR}} count(inventory.*)$<$ 3 \\ \hline
  \end{tabular}
  }
  \caption{An example modified from that in Figure 5.}

  \label{figure2union}
  \end{table}

\section{Further Discussion on Executable SQL Generation}
In Section 3.6, we discuss that different questions in Figure 5 will be converted to different NatSQL, where training data is the key.
Firstly, in the dataset, for SQL with multiple  \textit{WHERE} conditions, the order of the conditions is mostly consistent with the question. 
Secondly, the NatSQL further expands this type of training data. 
For example, the NatSQL queries in Figure 1,3,4 contain more \textit{WHERE} conditions than SQL and other IRs, and these conditions appear in the order they are mentioned.
These training data make it possible for models to generate different NatSQL according to the different questions in Figure 5. 

\section{Gold NatSQL Error Analysis}

Table \ref{table4} presents the F1 score of NatSQL for different SQL components. We observe that the main errors come from \textit{GROUP BY} and IUE matching. Although NatSQL cannot be converted to all gold \textit{GROUP BY} clauses, most of these errors don't affect the execution results. The IUE errors occur because NatSQL only supports one IUE operator per query.

\begin{table}[h]
    \centering
    \resizebox{.99\columnwidth}{!}{
    \smallskip\begin{tabular}{cccc}
        \hline
        Component&F1 &Component&F1 \\
        \hline \hline
        select & 0.997 & where & 0.969 \\ 
        group & 0.879 & order & 0.996 \\ 
        and/or & 0.998 & IUE & 0.900  \\
        keywords  & 0.989&& \\      
        \hline
    \end{tabular}
    }
    \caption{Partial matching F1 score of NatSQL on the Spider development set.}\smallskip

    \label{table4}
\end{table} 

Some other errors are due to the limitation of the exact match evaluation method when evaluating the \textit{JOIN ON} clause of subqueries and sub-subqueries. Specifically, when the \textit{FROM} and \textit{JOIN} in a generated subquery is not identical to the gold SQL, the Spider evaluation scheme considers it to be wrong. For example, the following two SQL statements have the same semantic meaning, but they are recognized as different by the Spider exact match evaluation method, thus results in an exact match error. \\
... \ col in ( {\bf SELECT} col {\bf FROM} T1 {\bf JOIN} T2 ... )\\
... \ col in ( {\bf SELECT} col {\bf FROM} T2 {\bf JOIN} T1 ... )\\


\section{SQL, SemQL and NatSQL Examples}
We present more examples in Table~\ref{tab:ex-app}.
\begin{table*}[h!]
    \centering
    \begin{tabular}{p{0.14\columnwidth}p{1.8\columnwidth}}

    Ques:& What are the name of the countries where there is not a single car maker? \\\hline
  
  SQL: & {\bf SELECT} CountryName {\bf FROM} countries  {\bf EXCEPT}  {\bf SELECT} T1.CountryName {\bf FROM} countries AS T1 {\bf JOIN} car\_makers AS T2 {\bf ON} T1.countryId  =  T2.Country;
  \\\hline
  SemQL: & Not Support \\\hline
  
  NatSQL:    &  {\bf SELECT} countries.countryname {\bf WHERE} {\bf except} @ is car\_makers.* \\\hline
    \\
  \\
    Ques: & Find the last name of the students who currently live in the state of North Carolina but have not registered in any degree program. \\\hline

    SQL: & 
    {\bf SELECT} T1.staff\_name  {\bf FROM} staff AS T1 {\bf JOIN} Staff\_DA AS T2 {\bf ON} T1.staff\_id  =  T2.staff\_id {\bf WHERE} T2.job\_title\_code  =  "Sales Person"  {\bf EXCEPT} \\ &
    {\bf SELECT} T1.staff\_name  {\bf FROM} staff AS T1 {\bf JOIN} Staff\_DA AS T2 {\bf ON} T1.staff\_id  =  T2.staff\_id {\bf WHERE} T2.job\_title\_code  =  "Clerical Staff" \\
    \hline
    SemQL: & 
    {\bf SELECT} staff.staff\_name  {\bf WHERE} Staff\_DA.job\_title\_code  =  "Sales Person"  {\bf EXCEPT}  \\ &  
    {\bf SELECT} staff.staff\_name  {\bf WHERE} Staff\_DA.job\_title\_code  =  "Clerical Staff"  \\
    \hline
  
    NatSQL: & 
    {\bf SELECT} staff.staff\_name  {\bf WHERE} Staff\_DA.job\_title\_code  =  "Sales Person" \\ & {\bf AND} Staff\_DA.job\_title\_code  !=  "Clerical Staff" \\
    
    \hline
    \\

    Ques: & Find id of the tv channels that from the countries where have more than two tv channels. \\\hline

    SQL: & 
    {\bf SELECT} id {\bf FROM} tv\_channel {\bf GROUP BY}   country {\bf HAVING} count(*)  $>$  2 \\
    \hline
    SemQL: & 
    {\bf SELECT} tv\_channel.id  {\bf WHERE} count ( tv\_channel.* ) $>$ 2 \\
    \hline
  
    NatSQL: & 
    {\bf SELECT} tv\_channel.id  {\bf WHERE} count ( tv\_channel.* ) $>$ 2 \\
    
    \hline

    \\
  
    Ques: & List all song names by singers above the average age. \\\hline

    SQL: & 
    {\bf SELECT} song\_name {\bf FROM} singer {\bf WHERE} age  $>$  ( {\bf SELECT} avg(age) {\bf FROM} singer ) \\
    \hline
    SemQL: & 
    {\bf SELECT} singer.song\_name {\bf WHERE} singer.age $>$ ( {\bf SELECT} avg(singer.age) ) \\
    \hline
  
    NatSQL: & 
    {\bf SELECT} singer.song\_name singer {\bf WHERE} @ $>$ avg ( age )\\
    \hline

    \\
  
    Ques: & Which district has both stores with less than 3000 products and stores with more than 10000 products? \\\hline

    SQL: & 
    {\bf SELECT} district {\bf FROM} shop {\bf WHERE} Number\_products $<$  3000 {\bf INTERSECT} {\bf SELECT} district {\bf FROM} shop {\bf WHERE} Number\_products $>$ 10000\\
    \hline
    SemQL: & 
    {\bf SELECT} shop.district {\bf WHERE} shop.Number\_products  $<$  3000 {\bf INTERSECT} {\bf SELECT} shop.district  {\bf WHERE} shop.Number\_products $>$  10000 \\
    \hline
  
    NatSQL: & 
    {\bf SELECT} shop.district {\bf WHERE} shop.number\_products $<$ 3000 {\bf and} shop.number\_products $>$ 10000\\

    \hline
    \end{tabular}
    \caption{SQL, SemQL and NatSQL examples from the Spider.}   
    \label{tab:ex-app}
    \end{table*}



\end{document}